\documentclass[11pt,a4paper]{article}
\usepackage[nohyperref]{emnlp-ijcnlp-2019}
\usepackage{times}
\usepackage{latexsym}
\usepackage{amsmath}
\usepackage{graphicx}
\usepackage{subfigure}
\usepackage{url}
\usepackage{siunitx} 
\usepackage{booktabs} 
\usepackage{multirow}
\usepackage{graphics}
\usepackage{multicol}
\usepackage{color, colortbl}
\usepackage{cleveref}

\aclfinalcopy 

\title{Multi-Image Summarization: Textual Summary from a Set of Coherent Images}

\author{
Nicholas Trieu, Sebastian Goodman, Pradyumna Narayana, Kazoo Sone, Radu Soricut\\
Google\\
{\tt \{ntrieu,seabass,pradyn,sone,rsoricut\}@google.com}
  }

\date{}

\begin{document}
\maketitle
\begin{abstract}
Multi-sentence summarization is a well studied problem in NLP, while generating image descriptions for a single image is a well studied problem in Computer Vision. However, for applications such as image cluster labeling or web page summarization, summarizing a set of  images is also a useful and challenging task. This paper proposes the new task of multi-image summarization, which aims to generate  a concise and descriptive textual summary given a coherent set of input images.
We propose a model that extends the image-captioning Transformer-based architecture for single image to multi-image. A  dense  average  image  feature  aggregation network allows  the  model to  focus  on a coherent  subset  of  attributes  across the input  images. We explore various input representations to the Transformer network and empirically show that aggregated image features are superior to individual image embeddings. We additionally show that the performance of the model is further improved by pretraining the model parameters on a single-image captioning  task, which appears to be particularly effective in  eliminating  hallucinations in the output.
\end{abstract}

\section{Introduction}

There has been an large amount of work on abstractive summarization (generating a textual summary from a text document or multiple documents) over the last years~\cite{rush2015neural,luong2015multi,nallapati2016abstractive,see2017get,paulus2017deep,amplayo2018entity,gehrmann2018bottom}.
 However, there are other cases for which abstractively producing a textual summary over a set of inputs is useful. For example, consider the task of providing a textual description for image clusters. Given multiple images of dogs, such as Collie, German Shepherd and Australian Shepherd, we could describe these as images of ``Herding Dogs'' or just ``Dogs''. At the same time, if the input cluster contains other types of animals such as tiger or elephant, we could say these are ``Mammals'' or ``Animals''. In other words, given a set of $N (> 1)$ images that share some common attributes, the task is to generate a concise yet most specific \& descriptive text that is applicable to \textit{all} the input images. We call this task the multi-image summarization problem. The challenge of the multi-image summarization problem is to find the right level of ``abstraction'' to describe the given image set. In the earlier examples, depending on the other images in the set, the same Collie image could be part of a set labeled as either ``Herding Dog'', ``Dog'', or ``Animal''; while technically the last two are still correct even when all the images are about shepherd dogs, we usually consider the first to be the correct level of abstraction and thus desirable.


Our approach to the multi-image summarization task is to train an end-to-end model that directly generates the summary given the set of images. In this approach, the main challenge for a model is to identify the common attributes that are present in all input images. 
%
We note here that multi-image summarization may be seen as a generalization of the single-image captioning task, in that multi-image summarization needs to operate over features of not only a single image but multiple images, and it needs to find a unified representation over the input features from which the model can generate textual output.


One challenge for training such a multi-image summarization model is the availability of datasets that have captions for multiple images, as we are not aware of any usable dataset of this kind (see \citet{Hossain:2018} for a comprehensive review of image captioning datases and models and dataset, as it does not contain any dataset for sets of related images with textual annotations over a vocabulary that spans more than 100 words\footnote{The only exception is WordNet labels in ImageNet but their label vocabulary is too coarse and not flexible enough for the applications we target.}). There are two possible venues to investigate in this regard: build such a dataset from scratch, given that we can exploit existing web-page annotations to achieve our goal; and, devise a mechanism for taking advantage of single-image caption datasets and apply them towards the multi-image summarization task.

This paper has three primary contributions. First, we propose a new multi-image summarization task, accompanied by a detailed description of the process by which to construct a dataset that can be directly used to do supervised learning for the task. Second, we study how various representations and modeling decisions impact the performance of the resulting learned models. Third, we show how one can take advantage of single-image captioning annotations to further improve the performance on the multi-image summarization task.


\section{Related Work}

As briefly mentioned in the previous section, the present study is deeply related to multi-document text summarization as well as single-image captioning. This section covers related work from these two areas.


\subsection{Multi-Document Summarization}

One of the related tasks is a multi-document summarization, which is concerned about generating a short piece of summary text from a set of textual documents. Like multi-image summarization, multi-document summarization tries to summarize a group of related entities with a concise text description, where each entity itself could have an individual text description. Our approach in aggregating individual pre-trained image embeddings is similar to that of \citet{Mani:2018}, which uses the centroid vector of individual document vectors.
\citet{Cao:2017} have similarly applied pretrained features learned from an individual-text dataset to a multi-document dataset. One difference between their approach and ours is we do not need a separate model to learn from single-caption datasets. Others \cite[such as][]{Nayeem:2018}
apply a compositional approach to generate multiple fusions of individual outputs and learn a ranking of the fusions to select a final summary. This approach of aggregating caption outputs is not ideal for our case, as the generated captions already lose much information from the input image embeddings. However, we emulate the fusion and ranking via an attentional approach to aggregating the individual image embeddings.



\subsection{Single-Image Captioning}

Most deep learning architectures for single-image captioning use an encoder-decoder structure where a CNN encodes the raw image bytes, then some language model decoder such as an RNN decodes into a caption. As detailed by \citet{Hossain:2018}, more than 40 published image captioning models since 2015 use some form of CNN encoder and RNN decoder.
Recent advances in NLP have replaced the RNN with the Transformer~\cite{Vaswani:2017}
which dispenses with recurrence and uses solely an attention mechanism for sequence modeling. We follow recent image captioning work~\cite[such as][]{Sharma:2018, Zhu:2018, zhao2019informative} in using Transformer Networks to decode into captions.



\section{Multi-Image Summarization Dataset}

\subsection{Creation Process}

The process by which we construct our multi-image caption dataset consists of a Flume~\cite{Chambers:2010:FEE:1809028.1806638} pipeline to processes billions of web pages in parallel to find ones that contain the type of annotation we need. This is similar to (and inspired by) the work of \citet{Sharma:2018}, in which they used Alt-text annotations and the corresponding images from a large web corpus to create a single-image caption dataset.


Our data source is a subset of the Web pages that meet the following criteria.
First,  we retain the web pages that a document content classification model available from the Google Cloud API~\cite{gcp_text_classification_api} identifies as ``Shopping''. Web pages in the shopping category tend to have various products of a similar type that share multiple attributes~\cite[e.g.][]{Han:2017}, making them good candidates for this multi-image summarization task.
Second, we use the single-product vs. multi-product page type classifier described by \citet{Vovk:2019} to retain only the pages that are identified as multi-product.
The underlying assumption here is that a multi-product page contains a variety of images for different products that also share a common ``theme''; from this point on, we extract the images from these pages and refer to them as ``image groups''.

\begin{figure}[!h]
\centering
\includegraphics[width=.9\linewidth]{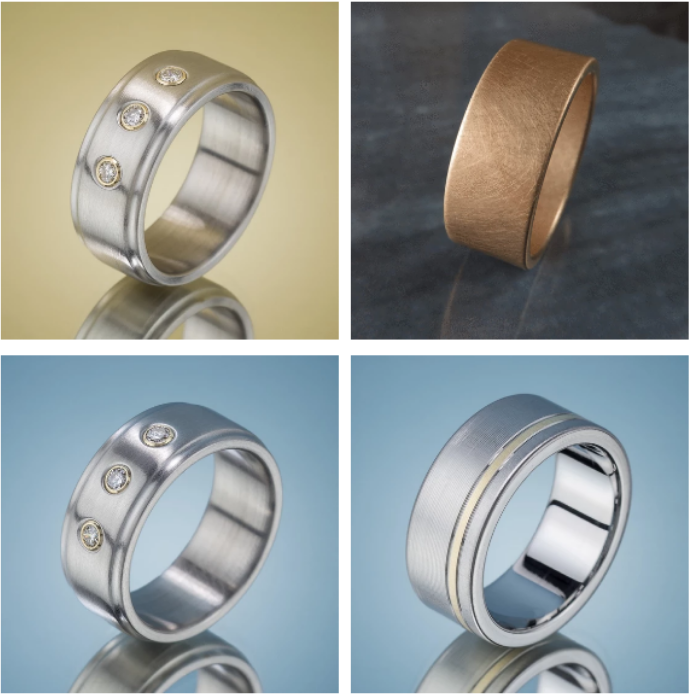}
\centering
\caption{A typical example. Groundtruth is ``wedding bands''. See Table~\ref{tab:caption_list} for generated summaries.}
\label{figure:example_data_bands}
\end{figure}

In the next steps, the pipeline only keeps the image groups that contain at least 5 images where all of the images satisfy the following conditions:
the images are of JPEG format where both dimensions are greater than 100 pixels; the aspect ratio of each image is at most 3; 
not trigger pornography or profanity detectors (from the Google Cloud API).

Next, the logic of the pipeline focuses on extracting the groundtruth labels for the image groups.
For the experiments presented here, the pipeline only considers pages identified as containing text in English, and it selects the page's title as the candidate for the groundtruth label.
The candidate is further processed with respects to mentions of brands (e.g., Nike) and specific product names (e.g., Air Jordan).
Fine-grained entity labels such as brand and product names are numerous, sparse, and therefore difficult to learn.
We decided to remove such entity labels from the candidate groundtruth, in order to decouple the multi-image captioning task from the task of fine-grained entities recognition.
To remove the occurrence of these entity labels, we first locate all instances of certain entity
types\footnote{The types are: Organizations, Business Operations, Inventors, Employers, People, Automobile Models, Product Lines, Websites, TV/Movie series, Company, Locations, Events, Fashion labels, Sports teams.}
using a named entity recognizer, and then delete the corresponding characters.

In a final step that ensures that the candidate groundtruth matches the semantic content of its image group, the pipeline filters out candidates for which none of the groundtruth text tokens can be mapped to Google Cloud Vision API object identification labels returned over the set of images in the image group.

\begin{figure}%
\centering
\includegraphics[width=.9\linewidth]{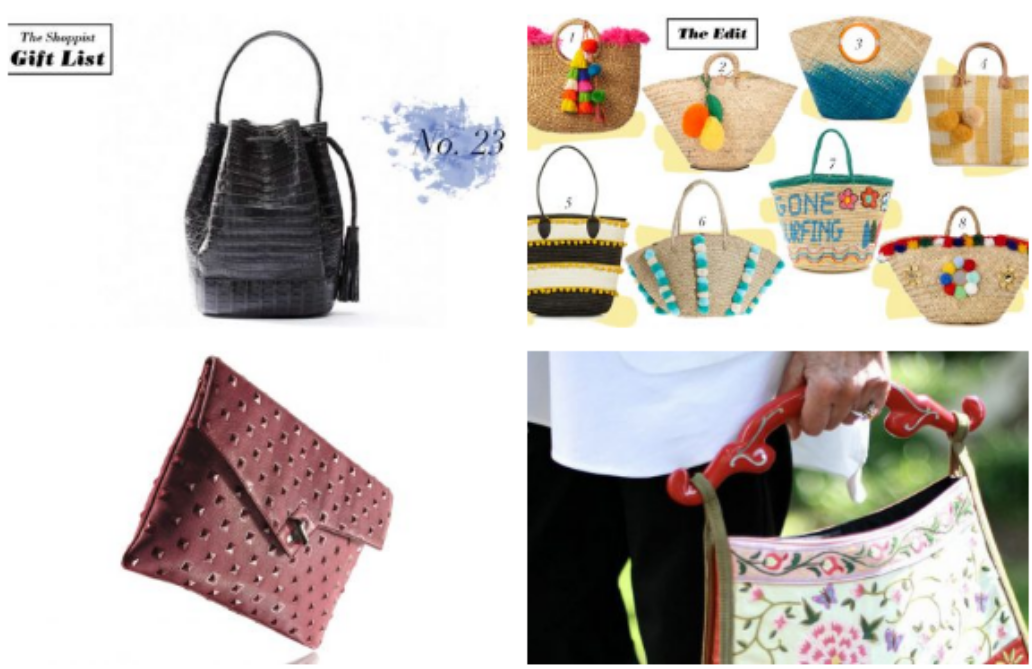}
\centering
\caption{Groundtruth is ``bags'' with 
very different style of images.
See Table~\ref{tab:caption_list} for generated summaries by various models.}
\label{figure:example_data_bag}
\end{figure}

\begin{figure}%
\centering
\includegraphics[width=.9\linewidth]{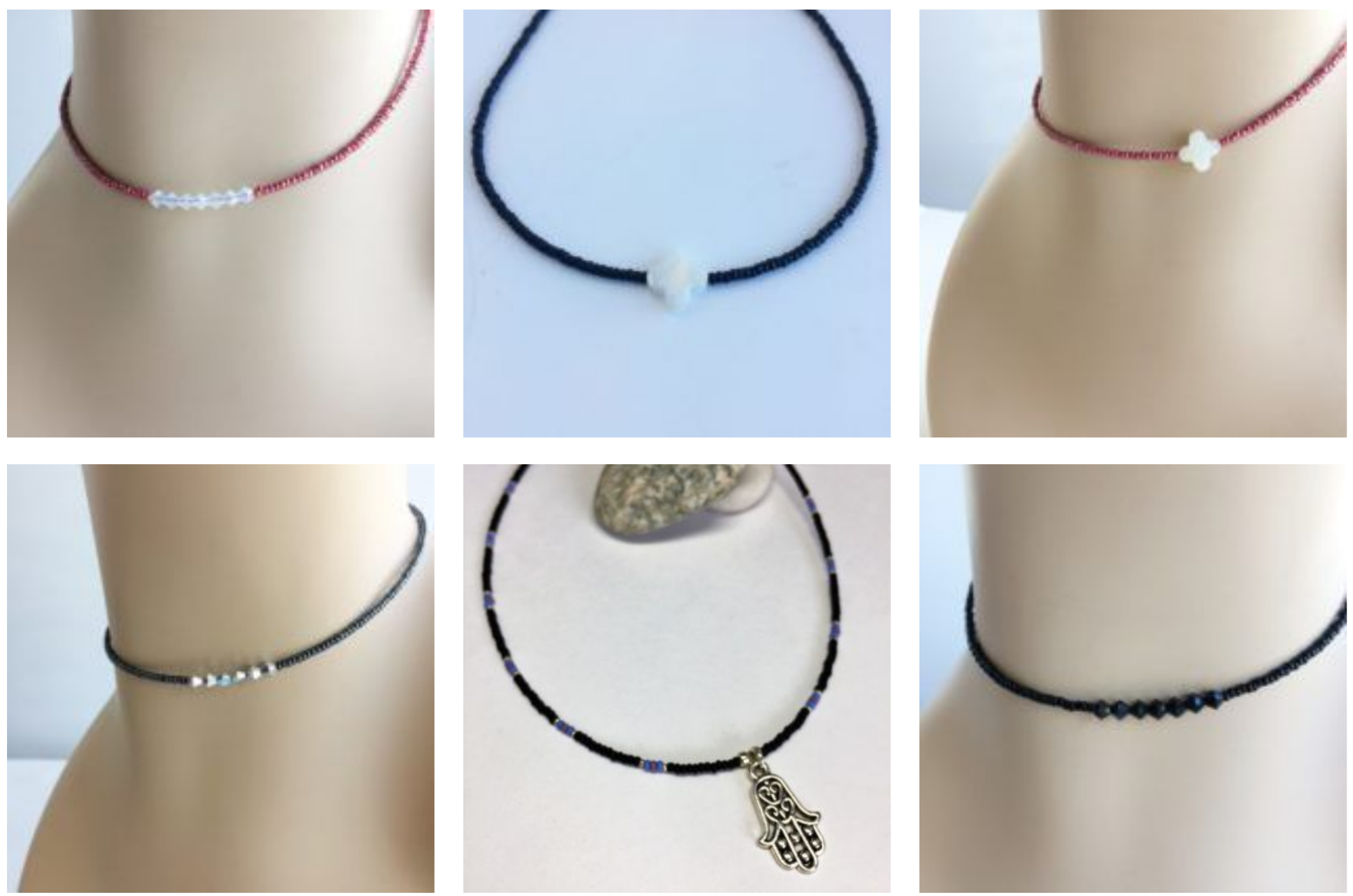}
\centering
\caption{Groundtruth is ``beaded choker'' (The set contains 2 more similar images not shown here). See Table~\ref{tab:caption_list} for generated summaries by various models.}
\label{figure:example_data_choker}
\end{figure}

We illustrate the resulting examples in ~\cref{figure:example_data_bands,figure:example_data_bag,figure:example_data_copper,figure:example_data_choker}, which show several instances of the examples extracted by our pipeline that pass all the filters described above.

\begin{table*}[]
\caption{Multi-Image caption dataset statistics for train and dev+test sets. First row corresponds to train set stats and second row corresponds to 5000 validation examples and 80,816 test examples.}
\label{tab:data_stats}
\resizebox{\textwidth}{!}{
\begin{tabular}{@{}cccccccc@{}}
\toprule
      & Size & \begin{tabular}[c]{@{}c@{}}Average \\ Aspect Ratio\end{tabular} & \begin{tabular}[c]{@{}c@{}}Average\\ Image Height\end{tabular} & \begin{tabular}[c]{@{}c@{}}Average\\ Image Width\end{tabular} & \begin{tabular}[c]{@{}c@{}}Average\\ Caption Bytes\end{tabular} & \begin{tabular}[c]{@{}c@{}}Average\\ Caption Tokens\end{tabular} & \begin{tabular}[c]{@{}c@{}}Average\\ Num Images\end{tabular} \\ \midrule
Train & 2,018,239 & 0.957                                                           & 542                                                            & 494                                                           & 24.53                                                               & 3.85                                                                 & 16.80                                                        \\
Valid+Test   & 85,816    & 0.915                                                           & 534                                                            & 486                                                           & 21.64                                                               & 3.33                                                                 & 16.88                  
                                                     \\ \bottomrule
\end{tabular}
}
\end{table*}

\subsection{Dataset Stats}

Running the pipeline described above results in about 2.1 Million examples.
We hold out 5,000 for validation, 80,816 for test, and use the remaining as training data.
A tokenized version of the groundtruth label results in a total vocabulary size of 197,745 distinct token types over the training set.
Table~\ref{tab:data_stats} provides additional dataset statistics such as average aspect ratio, average caption length in bytes, average tokens per groundtruth label, etc.
For the purpose of training our models over this dataset, the groundtruth labels are tokenized by a Wordpiece model~\cite{Schuster:2012}, using the same algorithm as the one in~\cite{Devlin:2018}.

\section{Multi-image Summarization Models}

\subsection{Image Features}
For each of the images in our dataset, we use pretrained individual image embeddings using Graph-RISE embeddings~\cite{Juan:2019}.
Graph-RISE learned embeddings (for single images) are trained using a ResNet neural network architecture with about 40 Million semantic labels. Note that there is no inherent clustering or embedding aggregation enforced by Graph-RISE. However, CNN-based features have been shown to cluster well for unsupervised tasks~\cite{Guerin:2018}, which aligns well with our objective.

\subsection{Models}

\begin{figure}%
\centering
\includegraphics[width=1.0\linewidth]{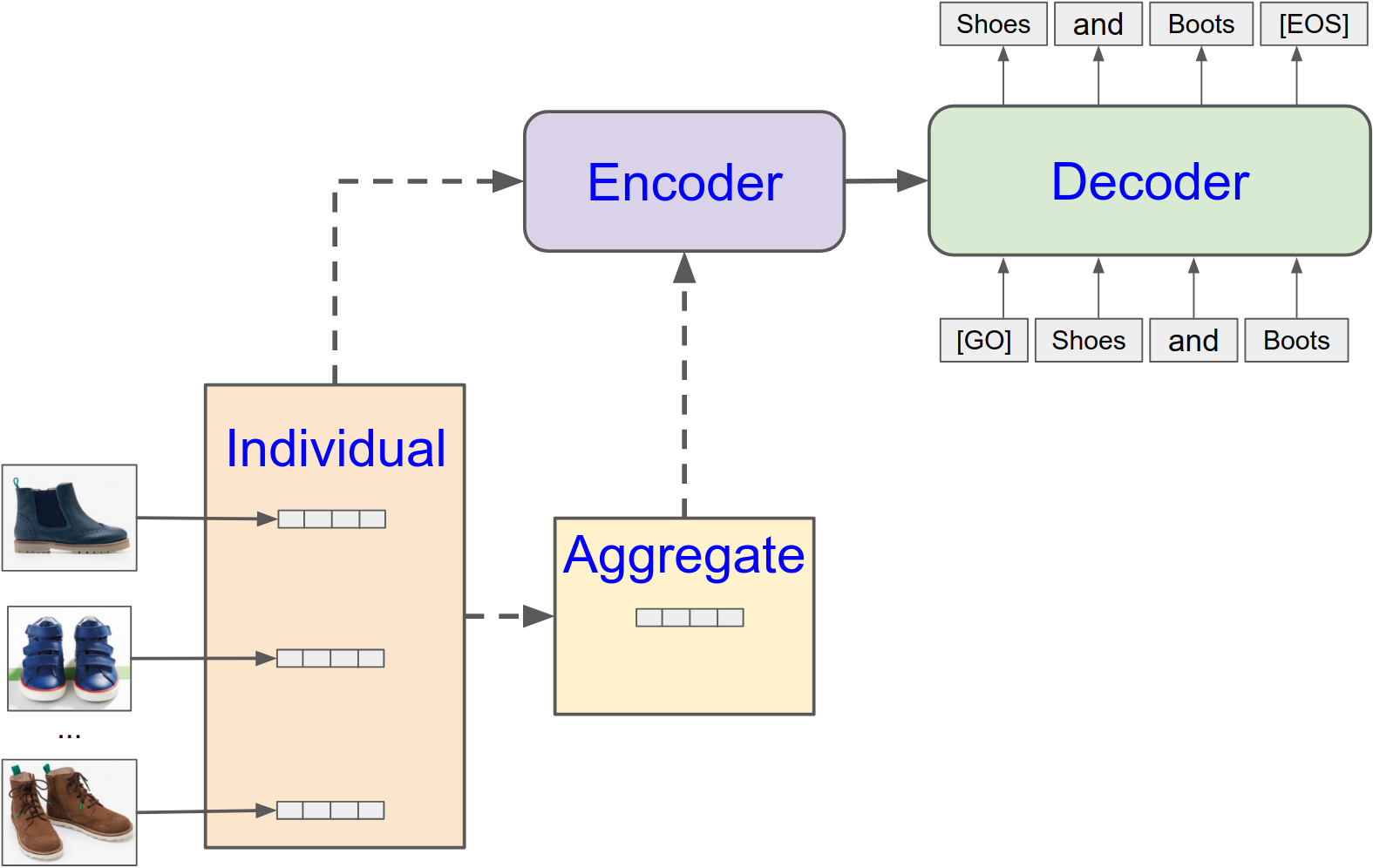}
\centering
\caption{Multi-image Model encoder-decoder structure. The encoder may optionally take in either the individual or aggregate image embeddings.}
\label{figure:schematic}
\end{figure}

Our models follow the encoder-decoder structure shown in Figure~\ref{figure:schematic}. There are four main components to each model:

\begin{itemize}
    \item An externally trained image embeddings for each image (Graph-RISE), frozen during training.
    \item An optional aggregation step that outputs a vector of aggregated image embeddings for the image group.
    \item A BERT~\cite{Devlin:2018} Encoder module that may take in (1) the individual image embeddings, or (2) the aggregated image embeddings (or both), and applies self-attention across these features to encode them into a tensor H.
    \item A BERT Decoder module that generates outputs conditioned on H and attentional inputs.
\end{itemize}

Our hypothesis is that if the pretrained image embeddings are trained in such a way that similar concepts are clustered together in the embedding space, we should be able to generate textual description most robustly from the centroid of that cluster. Then our task is equivalent of finding the centroid of each concept -- we call that point in the embedding space a ``canonical embedding'' for the given concept. Since all input images are expected to have a shared concept, the canonical embedding point is expected to lie within the minimum bounding box created by the set of points from each image embedding. Furthermore, if the concept occupies $k$-dimensional subspace of the entire $K$-dimensional embedding space (due to the way external embeddings was trained), distribution of embedding in each dimension should help our model identify those $k$-dimensional subspace.
Thus, an important component of our work is finding the best aggregation approach for use in our models. \\ 

Let $N$ be the number of images in the group, where each image embedding, $\mathbf{e}$, is a $K$-dimensional float vector ($K=64$ in this study), and let $\mathrm{E}$ denote the individual image embeddings matrix of shape $(N, 64)$, $\mathbf{E} = \big[\mathbf{e}^{
        (0)} \vert \mathbf{e}^{(1)} \vert \cdots  \vert  \mathbf{e}^{(N-1)} \big]^T$, where the vertical bar $\vert$ denotes row-wise concatenation of column vectors. In other words, $E_{ij} = e^{(i)}_j$.
We investigate five different approaches.

\begin{figure}%
\centering
\includegraphics[width=.9\linewidth]{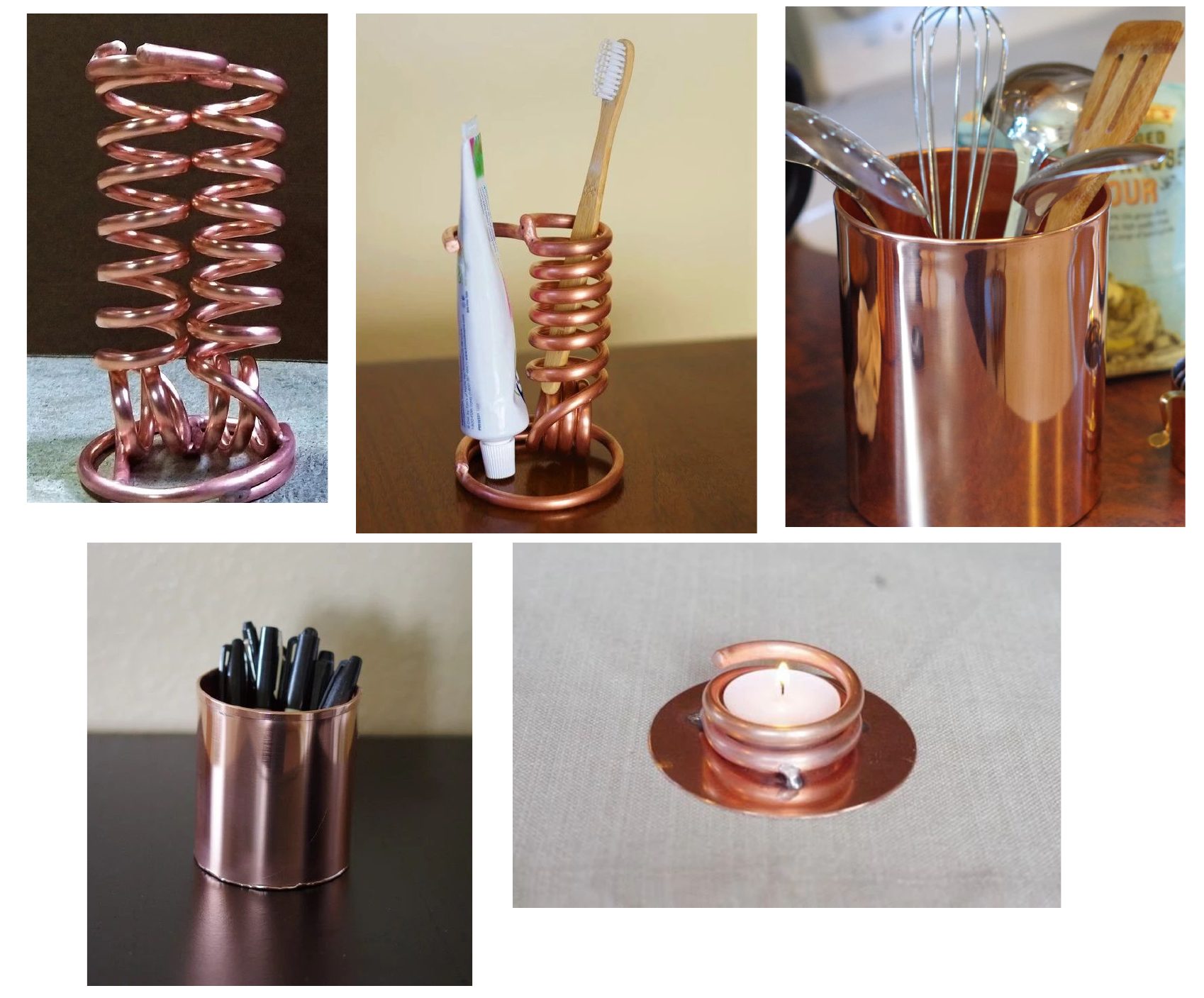}
\centering
\caption{Groundtruth is ``copper gadgets''.
This is a hard case as what is common is the material.
See Table~\ref{tab:caption_list} for generated summaries by various models.
}
\label{figure:example_data_copper}
\end{figure}

\begin{enumerate}
    \item No aggregation, i.e. only individual image embeddings $\mathbf{e}^{(
    i)}$ 
    are fed into the encoder.
    
    \item Element-wise standard deviation $\mathbf{\sigma}$ of the image embeddings. Given $N$ images, for each of the $K$ embedding dimensions, we take the standard deviation across the $N$ images to get a $K$-dimension standard deviation vector. Mathematically, $\mathbf{\sigma}$ is defined as
    \begin{align*}
     \sigma_j &= \sqrt{\frac{1}{N}\sum_{i} (e_{j}^{(i)}-\overline{e}_{j})^2}
    \end{align*}
    where $e_j^{(i)}$ represents $j$-th element of the $i$-th image embedding vector, and $\overline{e}_{j} = {1}/{N}\sum_{i} e_{j}^{(i)}$ is an element-wise average of all image embedding vectors.\\
    Standard deviation of the image embeddings are always fed into the captioning model in combination with other (single or aggregated) image embeddings. 
    The intuition behind this is that, for example, if the ``copper'' attribute (see Fig.~\ref{figure:example_data_copper}) is represented at locations, say $i\ldots i+j$ for image 1, then it is represented at the same locations for all the other image representations in the image group; therefore, the standard deviation feature will ensure that the resulting vector has a low magnitude for shared concepts, and high magnitude for non-shared concepts. This signal can be successfully exploited by our models, as our experimental results will indicate.
    
    
    \item Fixed weight averaging ($a^F$) of the image embeddings. Given $N$ images each with a $K$-dimension embedding, we average across the $N$ images to get a $K$-dimension average embedding. Each image has a fixed averaging weight of $1/N$. Mathematically, $a^F$ is defined as
    \[
    a^F_j = \overline{e}_{j}. 
    \]
    
    \item Dense weighted averaging $a^D$ of the image embeddings.
    Instead of using a fixed uniform weights across all images, we learn the averaging weights via an intermediate dense layer.
    We compute $a^D$ as follows. First create $64\times N$-dimensional column vector by taking concatenation of all image embeddings, 
    \begin{align*}
        \mathbf{e}^{f} &= \big[\mathbf{e}^{
        (0)T} \vert \mathbf{e}^{(1)T} \vert \cdots  \vert  \mathbf{e}^{(N-1)T} \big]^T.
    \end{align*}
    Let $H$ be a learned hidden matrix of shape $(N, K\times N)$ and using $e^{f}$, our image-wise aggregation weights are
    \begin{align*}
        a^D_j &= \sum_{i=0}^{N-1} w_i e_j^{(i)},
    \end{align*}
    where $w$ is an $N$-dimensional vector, $w=\text{softmax}(\mathbf{H}\mathbf{e}^{f})$.
    
    
    \item Self-attention weighted averaging $a^S$ of the image embeddings. We learn the averaging weights across the images via an intermediate self-attention mechanism just like the dense layer approach except this time we use self-attention. Specifically, we use scaled dot product attention.\\ 
    Let $\mathbf{H}_1$ and $\mathbf{H}_2$ be learned hidden matrices of shape $(K, M)$, where $M$ is the desired aggregation embedding dimension. With $\mathbf{Q} = \mathbf{E}\mathbf{H}_1$, $\mathbf{K} = \mathbf{E}\mathbf{H}_2$ and $\mathbf{V} = \mathbf{E}$ as query, key and value matrix of shape $(N, M)$, $(N, M)$ and $(N, K)$ respectively, we compute $a^S$ as follows:
    \begin{align*}
        \mathbf{W} &= \text{softmax}\left(\frac{\mathbf{Q}\mathbf{K}^T}{\sqrt{M}}\right)\mathbf{V}, \text{and} \\
        a^S_j &= \frac{1}{N}\sum_{i=0}^{N-1} W_{ij} E_{ij}.
    \end{align*}
    
\end{enumerate}

These approaches add increasing complexity to how the model aggregates the individual image embeddings. Calculating the fixed weighted averaging before feeding it into the encoder helps prevent the model from overfitting on individual image embeddings. Using a dense layer to learn the averaging weights allows the model to put emphasis on certain images or ignore outliers. Replacing the dense layer with a self-attention mechanism makes the model order-invariant with respect to the input images.

\section{Experiments}

We evaluate the effect of each aggregation approach on the model performance for our multi-image summarization task. 

\subsection{Experimental Setup}
We perform additional processing of the data for our experiments, as follows.
Training captions are truncated to a maximum of 32 (word-piece) tokens. Each token must appear at least 10 times in the dataset in order to be a part of the model; all other tokens are replaced with a special token $\langle$UNK$\rangle$ throughout this study.

All models are trained using MLE loss and optimized using Adam~\cite{kingma2014adam} with learning rate $2\times 10^{-5}$ and  batch size of 1024. Models are trained until their maximum validation CIDEr~\cite{vedantam2015cider} score does not increase for 100K steps. Checkpoints are saved when a model reaches a new maximum validation CIDEr score. The final model is selected based on the best performing checkpoint on the validation set.
We use beam search with a beam size of 4 during decoding.

\subsection{Comparative Performance}

\begin{table}[ht!]
\centering
\caption{Scores for different model inputs.
The ``Indiv'' column denotes whether the model uses the individual image embeddings in addition to any possible aggregation features. The $\mathbf{\sigma}$ column denotes whether the model uses the standard deviation. The reported test set scores correspond to  the best validation model.
}
\label{tab:main_results_nostdev}
\resizebox{\columnwidth}{!}{
\begin{tabular}{@{}cccccc@{}}
\toprule
Indiv & Averaging & $\mathbf{\sigma}$  & CIDEr & BLEU-4 & ROUGE-L \\ \midrule
Y     &            &           & 1.331         & 0.244           & 0.423      \\ 
Y     & Fixed &                & 1.307         &  0.242          & 0.422     \\ 
Y     & Dense &               & 1.312         & 0.242           & 0.421     \\
Y     & Self-Attn &            & 1.313     & 0.243           & 0.424      \\ 
      & Fixed &               & 1.342          & 0.244           & \textbf{0.430}     \\ 
      & Dense  &            & \textbf{1.362}& \textbf{0.245} & \textbf{0.430}      \\ 
      & Self-Attn &        & 1.356    & \textbf{0.245}           & \textbf{0.430}   \\ 
\end{tabular}
}


\resizebox{\columnwidth}{!}{
\begin{tabular}{@{}cccccc@{}}
\toprule
Indiv & Averaging & $\mathbf{\sigma}$ & CIDEr & BLEU-4 & ROUGE-L \\ \midrule
Y     &           & Y              & 1.295              & 0.241            & 0.415     \\ 
Y     & Fixed & Y             & 1.267              & 0.238           & 0.408     \\ 
Y     & Dense  & Y             & 1.251              & 0.236           & 0.407     \\
Y     & Self-Attn & Y             & 1.241              & 0.237           & 0.404     \\ 
      & Fixed & Y             & 1.361                 & 0.243        &  0.416    \\ 
      & Dense  & Y             & \textbf{1.382}    & \textbf{0.247}   & \textbf{0.431}       \\
      & Self-Attn & Y      & 1.373         & 0.246                     & 0.425       \\  \bottomrule

\end{tabular}
}
\end{table}


Table~\ref{tab:main_results_nostdev} shows scores for models trained on different input combinations using the standard automated evaluation metrics CIDEr~\cite{vedantam2015cider}, BLEU~\cite{papineni2002bleu} and ROUGE~\cite{lin2004rouge}). Each row represents a different model, where a non-empty cell at an input feature column signifies that the model was trained with that input.
Both model-learned weighted averaging aggregation approaches (Self-Attn and Dense) have achieved some of the best performance across all automated evaluation metrics.
All forms of averaging (Fixed, Self-Attn and Dense) outperform models with individual image embedding features. It is worth noting that, despite the fact that aggregated features work well by themselves, when combined with individual image embedding features, the aggregated features consistently make the model perform worse than just using individual image embeddings.

When an element-wise standard deviation of embedding vector $\mathbf{\sigma}$ is added to each of these models, this difference becomes more pronounced; the models with aggregated features alone consistently outperformed the same model without the standard deviation feature, while the performance of the models with individual image embeddings (regardless of presence of aggregated embeddings features) significantly decreases. The former is in line with our expectation that there is canonical embedding subspace that is easier to identify with element-wise standard deviation features. The latter part indicates an inability of the models to learn to correctly downgrade, using the attention mechanisms, the contributions of the individual image embeddings; further investigation is needed to understand and correct this undesired outcome.

Furthermore, there is a consistent ordering of input performance with or without the standard deviation features. In order from best performing to worst, we have: dense weighted average, self-attention weighted average, fixed average, and finally individual embeddings. The ordering indicates a superiority of the dense weighted average approach for the multi-image summarization task. It also suggests there may be other features like the standard deviation that can help guide the weights of the averaging to obtain a canonical representation from multiple images.



\subsection{Image Ordering}

Recall that the non-aggregated feature vector contains the concatenation of the individual embeddings for each image. These embeddings are ordered in the vector according to their order on the webpage HTML text. It is therefore plausible that the order of the images might affect the model performance. We investigate this by randomly changing the image order for a subset of 500 test examples, then comparing the inference results against the original ordering.

Table~\ref{tab:order_ablation} shows the results for each ordering. As the numbers indicate, for some models the  performance gets worse while for others it improves. The small magnitude of the differences suggests that the models are not relying heavily on image order.
As expected, the scores for the fixed average model do not change at all as it is invariant to the image ordering. We observe small score differences even for the models that do not pass the non-aggregated feature vector to the encoder. This may be because the entire feature vector is mapped to some aggregate space to compute the attention scores, and changes to the image ordering may slightly affect the corresponding mapping.

\begin{table}[ht!]
\caption{CIDEr scores for different input image orderings on a sample of 500 test examples.
} \label{tab:order_ablation}
\resizebox{\columnwidth}{!}{
\begin{tabular}{@{}ccccc@{}}
\toprule
\multirow{2}{*}{Indiv} & \multirow{2}{*}{Averaging} & \multirow{2}{*}{$\mathrm{\sigma}$} & \multicolumn{2}{c}{CIDEr}   \\ \cmidrule(l){4-5} 
                       &                            &                    & Original order & Randomized \\ \cmidrule(r){1-5}
Y     &            &       & 1.954                 & 1.970             \\ 
Y     & Fixed  & Y     & 1.925                 & 1.925             \\ 
Y     & Dense  & Y     & 1.952                 & 1.947             \\ 
Y     & Self-Attn  & Y     & 1.954                 & 1.861             \\ 
      & Fixed  & Y     & 1.956                 & 1.956             \\ 
      & Dense  &  Y     & 1.990                   & 2.013             \\
      & Self-Attn   & Y     & 2.030                 & 2.046             \\ \bottomrule
\end{tabular}
}
\end{table}

\subsection{Single-Image Caption Pretraining}

\begin{table}[ht!]
\caption{CIDEr scores for pretrained vs non-pretrained models. Improvements can be observed across all types of models. Fixed Avg achieved the best performance after pretraining while Dense Avg is almost as good.
} \label{tab:pretrain}
\resizebox{\columnwidth}{!}{
\begin{tabular}{@{}ccccc@{}}
\toprule
\multirow{2}{*}{Indiv} & \multirow{2}{*}{Averaging} & \multirow{2}{*}{$\mathrm{\sigma}$} & \multicolumn{2}{c}{CIDEr}   \\ \cmidrule(l){4-5} 
                       &                            &                    & Not Pretrained & Pretrained \\ \cmidrule(r){1-5}
Y     &            & Y      & 1.331                & 1.406             \\ 
Y     & Fixed  & Y     &         1.267         & 1.372              \\ 
Y     & Dense  & Y     & 1.251                 & 1.345              \\
Y     & Self-Attn  & Y     &          1.241        & 1.378             \\ 
      & Fixed  & Y     & 1.361                 & \textbf{1.452}              \\ 
      & Dense  &  Y     & \textbf{1.382}                    & 1.443             \\
      & Self-Attn   & Y     & 1.37                 & 1.427              \\  \bottomrule
\end{tabular}
}
\end{table}

In this section, we investigate whether we can improve our model's performance by leveraging single-image caption datasets.
This is a difficult task for several reasons: (1) pretraining of individual feature models would be done on $(1, K)$-shaped single-image inputs, but finetuned on $(N, K)$-shaped multi-image inputs; (2) aggregate feature models would be ``aggregating'' just a single image during pretraining, potentially anti-learning how to properly aggregate. Despite these concerns, we show that pretraining yields significant improvements for both individual and aggregate feature models.

Using a Flume pipeline similar to the one used to create the multi-image caption dataset, we generate a single-image caption dataset from the same source and using the same generic filters. Instead of using $\langle$image group, page title$\rangle$ pairs, we instead generate $\langle$single image, caption $\rangle$ pairs using each image's Alt-text field as the caption, in a similar manner to \cite{Sharma:2018}. This approach results in about 100 million image and caption pairs.

We use this single-image caption dataset to pretrain both the BERT encoder and decoder for each model input configuration. First, we pretrain on the Alt-text single-image caption dataset using the same experimental setup as before. Next, a best single-image caption model is selected based on its performance on the multi-image caption validation set (adapted as a single-image dataset). We then train a new multi-image caption model on the multi-image caption dataset, initializing the BERT encoder and decoder weights with the values from the best single-image caption model for the corresponding input configuration.


Table~\ref{tab:pretrain} shows the results for each input configuration. Comparing pretrained vs non-pretrained models, we see that pretraining significantly helps all the model configurations. As before, all forms of averaging outperform models with individual image embeddings.

\begin{table*}[ht!]
\caption{Hand-picked examples and summaries.
In some instances, model outputs present ungrounded/hallucinated information such as ``genuine sapphire'' or ``human hair'' under the no-pretraining condition; these are corrected after pretraining.
In some other instances, more specific words such as ``choker'' or ``hoodies'' are generated after pretraining. All predictions shown are outputs for models using the standard deviation feature.}
\label{tab:caption_list}
\begin{tabular}{|l|p{185pt}|p{185pt}|}
\hline
Model & Not Pretrained & Pretrained \\ \hline
Groundtruth &
\multicolumn{2}{c|}{wedding bands (Fig.~\ref{figure:example_data_bands})} \\ \hline
Baseline &
rings & wedding rings \\ \hline
Fixed Avg &
men's wedding bands & rings \\ \hline
Dense Avg &
men's wedding bands & wedding bands \\ \hline
\hline

Groundtruth &
\multicolumn{2}{c|}{bags (Fig.~\ref{figure:example_data_bag})} \\ \hline
Baseline &
handbags & beaded leather bag \\ \hline
Fixed Avg &
handmade bag designs & beaded bag \\ \hline
Dense Avg &
handicraft bag & beaded bag \\ \hline
\hline

Groundtruth &
\multicolumn{2}{c|}{copper gadgets (Fig.~\ref{figure:example_data_copper})} \\ \hline
Baseline &
utensils & copper \\ \hline
Fixed Avg &
copper candlesticks & copper \\ \hline
 Dense Avg &
copper utensils & metal \\ \hline
\hline

 Groundtruth & \multicolumn{2}{c|}{ beaded choker (Fig.~\ref{figure:example_data_choker})} \\ \hline
Baseline &
necklace & necklaces \\ \hline
Fixed Avg & 
genuine sapphire bead choker necklace & beach bead choker \\ \hline
Dense Avg &
necklace &  choker necklaces \\ \hline
\hline

Groundtruth & 
\multicolumn{2}{c|}{synthetic lace wigs} \\ \hline
Baseline &
wigs & wigs \\ \hline
Fixed  Avg &
classic hair & wigs \\ \hline
Dense Avg &
human hair wigs & wigs \\ \hline
\hline

Groundtruth & \multicolumn{2}{c|}{sweatshirts \& hoodies} \\ \hline
Baseline &
hoodies hoodies \& sweatshirts & sweatshirts \\ \hline
Fixed Avg &
sweatshirts & sweatshirts \\ \hline
Dense Avg &
sweatshirts & hoodies \& sweatshirts \\ \hline

\end{tabular}

\end{table*}

Notably, under the pretraining condition, the fixed average model outperforms the self-attention average and dense average models. We hypothesize that this may be due to better pretraining on the Alt-text dataset, since these conditions use different pretrained weights.

We show in Table~\ref{tab:caption_list} several sampled summary outputs. Overall, we note that pretraining appears to help with eliminating unfounded (hallucinated) information that is not inferrable from the images, and also eliminate information that contradicts the semantic content in the images. Additionally, more specific attributes are correctly identified by the models trained under the pretraining regime.

\section{Conclusion}
In this paper, we describe a new task of multi-image summarization as the task of producing a concise and descriptive textual summary from a coherent set of images. We also describe a dense average image feature aggregation network that allows the model to focus on a coherent set of attributes common across the input images and improve performance as a result.

The model performance can be further improved by pretraining on a single-image captioning task, a result that indicates that single-image captioning datasets can be useful for multi-image summarization applications.

This is our first attempt at defining this task and experimenting with models designed to tackle it, and the results suggest that there is still room to further improve these models.

\bibliography{references}

\begin{thebibliography}{27}
\expandafter\ifx\csname natexlab\endcsname\relax\def\natexlab#1{#1}\fi

\bibitem[{Amplayo et~al.(2018)Amplayo, Lim, and Hwang}]{amplayo2018entity}
Reinald~Kim Amplayo, Seonjae Lim, and Seung-won Hwang. 2018.
\newblock Entity commonsense representation for neural abstractive
  summarization.
\newblock In \emph{Proceedings of NAACL-HLT 2018}, pages 697--707.

\bibitem[{Cao et~al.(2017)Cao, Li, Li, and Wei}]{Cao:2017}
Ziqiang Cao, Wenjie Li, Sujian Li, and Furu Wei. 2017.
\newblock \href
  {https://www.aaai.org/ocs/index.php/AAAI/AAAI17/paper/viewPaper/1452}
  {Improving multi-document summarization via text classification}.
\newblock In \emph{Proceedings of the Thirty-First AAAI Conference on
  Artificial Intelligence (AAAI-17)}.

\bibitem[{Chambers et~al.(2010)Chambers, Raniwala, Perry, Adams, Henry,
  Bradshaw, and Weizenbaum}]{Chambers:2010:FEE:1809028.1806638}
Craig Chambers, Ashish Raniwala, Frances Perry, Stephen Adams, Robert~R. Henry,
  Robert Bradshaw, and Nathan Weizenbaum. 2010.
\newblock \href {https://doi.org/10.1145/1809028.1806638} {Flumejava: Easy,
  efficient data-parallel pipelines}.
\newblock \emph{SIGPLAN Not.}, 45(6):363--375.

\bibitem[{Devlin et~al.(2018)Devlin, Chang, Lee, and Toutanova}]{Devlin:2018}
Jacob Devlin, Ming{-}Wei Chang, Kenton Lee, and Kristina Toutanova. 2018.
\newblock \href {http://arxiv.org/abs/1810.04805} {{BERT:} pre-training of deep
  bidirectional transformers for language understanding}.
\newblock \emph{CoRR}, abs/1810.04805.

\bibitem[{Gehrmann et~al.(2018)Gehrmann, Deng, and Rush}]{gehrmann2018bottom}
Sebastian Gehrmann, Yuntian Deng, and Alexander~M Rush. 2018.
\newblock Bottom-up abstractive summarization.
\newblock In \emph{Proceedings of the 2018 Conference on Empirical Methods in
  Natural Language Processing}.

\bibitem[{{Google Cloud}(2019)}]{gcp_text_classification_api}
{Google Cloud}. 2019.
\newblock \href
  {https://cloud.google.com/natural-language/docs/classifying-text}
  {{Classifying Content | Cloud Natural Language API | Google Cloud}}.

\bibitem[{Gu\'erin et~al.(2018)Gu\'erin, Gibaru, Thiery, and
  Nyiri}]{Guerin:2018}
Joris Gu\'erin, Olivier Gibaru, St\'ephane Thiery, and Eric Nyiri. 2018.
\newblock \href {https://arxiv.org/pdf/1707.01700.pdf} {{CNN features are also
  great at unsupervised classification}}.

\bibitem[{Han et~al.(2017)Han, Wu, Huang, Zhang, Zhu, Li, Zhao, and
  Davis}]{Han:2017}
Xintong Han, Zuxuan Wu, Phoenix~X Huang, Xiao Zhang, Menglong Zhu, Yuan Li,
  Yang Zhao, and Larry~S Davis. 2017.
\newblock \href
  {http://openaccess.thecvf.com/content_ICCV_2017/papers/Han_Automatic_Spatially-Aware_Fashion_ICCV_2017_paper.pdf}
  {Automatic spatially-aware fashion concept discovery}.

\bibitem[{Hossain et~al.(2018)Hossain, Sohel, Shiratuddin, and
  Laga}]{Hossain:2018}
Md.~Zakir Hossain, Ferdous Sohel, Mohd~Fairuz Shiratuddin, and Hamid Laga.
  2018.
\newblock \href {http://arxiv.org/abs/1810.04020} {A comprehensive survey of
  deep learning for image captioning}.
\newblock \emph{CoRR}, abs/1810.04020.

\bibitem[{Juan et~al.(2019)Juan, Lu, Li, Peng, Timofeev, Chen, Gao, Duerig,
  Tomkins, and Ravi}]{Juan:2019}
Da-Cheng Juan, Chun-Ta Lu, Zhen Li, Futang Peng, Aleksei Timofeev, Yi-Ting
  Chen, Yaxi Gao, Tom Duerig, Andrew Tomkins, and Sujith Ravi. 2019.
\newblock \href {http://arxiv.org/abs/1902.10814} {{Graph-RISE:
  Graph-Regularized Image Semantic Embedding}}.

\bibitem[{Kingma and Ba(2014)}]{kingma2014adam}
Diederik~P Kingma and Jimmy Ba. 2014.
\newblock Adam: A method for stochastic optimization.
\newblock \emph{arXiv preprint arXiv:1412.6980}.

\bibitem[{Lin(2004)}]{lin2004rouge}
Chin-Yew Lin. 2004.
\newblock Rouge: A package for automatic evaluation of summaries.
\newblock In \emph{Text summarization branches out}, pages 74--81.

\bibitem[{Luong et~al.(2015)Luong, Le, Sutskever, Vinyals, and
  Kaiser}]{luong2015multi}
Minh-Thang Luong, Quoc~V Le, Ilya Sutskever, Oriol Vinyals, and Lukasz Kaiser.
  2015.
\newblock Multi-task sequence to sequence learning.
\newblock \emph{arXiv preprint arXiv:1511.06114}.

\bibitem[{Mani et~al.(2018)Mani, Verma, Meisheri, and Dey}]{Mani:2018}
Kaustubh Mani, Ishan Verma, Hardik Meisheri, and Lipika Dey. 2018.
\newblock \href {http://arxiv.org/abs/1710.02745} {Multi-document summarization
  using distributed bag-of-words model}.

\bibitem[{Nallapati et~al.(2016)Nallapati, Zhou, Gulcehre, Xiang
  et~al.}]{nallapati2016abstractive}
Ramesh Nallapati, Bowen Zhou, Caglar Gulcehre, Bing Xiang, et~al. 2016.
\newblock Abstractive text summarization using sequence-to-sequence {RNN}s and
  beyond.
\newblock In \emph{Proceedings of CoNLL}.

\bibitem[{Nayeem et~al.(2018)Nayeem, Fuad, and Chali}]{Nayeem:2018}
Mir~Tafseer Nayeem, Tanvir~Ahmed Fuad, and Yllias Chali. 2018.
\newblock \href {https://www.aclweb.org/anthology/C18-1102} {Abstractive
  unsupervised multi-document summarization using paraphrastic sentence
  fusion}.
\newblock In \emph{Proceedings of the 27th International Conference on
  Computational Linguistics}.

\bibitem[{Papineni et~al.(2002)Papineni, Roukos, Ward, and
  Zhu}]{papineni2002bleu}
Kishore Papineni, Salim Roukos, Todd Ward, and Wei-Jing Zhu. 2002.
\newblock Bleu: a method for automatic evaluation of machine translation.
\newblock In \emph{Proceedings of the 40th annual meeting on association for
  computational linguistics}, pages 311--318. Association for Computational
  Linguistics.

\bibitem[{Paulus et~al.(2017)Paulus, Xiong, and Socher}]{paulus2017deep}
Romain Paulus, Caiming Xiong, and Richard Socher. 2017.
\newblock A deep reinforced model for abstractive summarization.
\newblock \emph{arXiv preprint arXiv:1705.04304}.

\bibitem[{Rush et~al.(2015)Rush, Chopra, and Weston}]{rush2015neural}
Alexander~M. Rush, Sumit Chopra, and Jason Weston. 2015.
\newblock A neural attention model for abstractive sentence summarization.
\newblock In \emph{Proceedings of EMNLP}, pages 379--389.

\bibitem[{Schuster and Nakajima(2012)}]{Schuster:2012}
M.~Schuster and K.~Nakajima. 2012.
\newblock Japanese and korean voice search.
\newblock \emph{{}IEEE International Conference on Acoustics, Speech and Signal
  Processing}.

\bibitem[{See et~al.(2017)See, Liu, and Manning}]{see2017get}
Abigail See, Peter~J. Liu, and Christopher~D. Manning. 2017.
\newblock Get to the point: Summarization with pointer-generator networks.
\newblock In \emph{Proceedings of the 55th Annual Meeting of the Association
  for Computational Linguistics}, pages 1073--1083.

\bibitem[{Sharma et~al.(2018)Sharma, Ding, Goodman, and Soricut}]{Sharma:2018}
Piyush Sharma, Nan Ding, Sebastian Goodman, and Radu Soricut. 2018.
\newblock \href {https://aclweb.org/anthology/P18-1238} {Conceptual captions: A
  cleaned, hypernymed, image alt-text dataset for automatic image captioning}.

\bibitem[{Vaswani et~al.(2017)Vaswani, Shazeer, Parmar, Uszkoreit, Jones,
  Gomez, Kaiser, and Polosukhin}]{Vaswani:2017}
Ashish Vaswani, Noam Shazeer, Niki Parmar, Jakob Uszkoreit, Llion Jones,
  Aidan~N. Gomez, Lukasz Kaiser, and Illia Polosukhin. 2017.
\newblock \href {https://arxiv.org/abs/1706.03762} {Attention is all you need}.

\bibitem[{Vedantam et~al.(2015)Vedantam, Lawrence~Zitnick, and
  Parikh}]{vedantam2015cider}
Ramakrishna Vedantam, C~Lawrence~Zitnick, and Devi Parikh. 2015.
\newblock Cider: Consensus-based image description evaluation.
\newblock In \emph{Proceedings of the IEEE conference on computer vision and
  pattern recognition}, pages 4566--4575.

\bibitem[{Vovk et~al.(2019)Vovk, Tochilkin, Narayana, Sone, and
  Basu}]{Vovk:2019}
Artem Vovk, Dmitrii Tochilkin, Pradyumna Narayana, Kazoo Sone, and Sugato Basu.
  2019.
\newblock Product phrase extraction from e-commerce pages.
\newblock In \emph{The Proceedings of The Web Conference 2019, Companion}.

\bibitem[{Zhao et~al.(2019)Zhao, Sharma, Levinboim, and
  Soricut}]{zhao2019informative}
Sanqiang Zhao, Piyush Sharma, Tomer Levinboim, and Radu Soricut. 2019.
\newblock Informative image captioning with external sources of information.
\newblock In \emph{Proceedings of ACL}.

\bibitem[{Zhu et~al.(2018)Zhu, Li, Liu, Peng, and Niu}]{Zhu:2018}
Xinxin Zhu, Lixiang Li, Jing Liu, Haipeng Peng, and Xinxin Niu. 2018.
\newblock \href {https://doi.org/10.3390/app8050739} {Captioning transformer
  with stacked attention modules}.
\newblock \emph{Applied Sciences}, 8(5).

\end{thebibliography}
\bibliographystyle{acl_natbib}

\end{document}